\documentclass[journal]{IEEEtran}

\usepackage{cite}
\ifCLASSINFOpdf
  \usepackage[pdftex]{graphicx}
\else
  \usepackage[dvips]{graphicx}
\fi
\usepackage[cmex10]{amsmath}
\usepackage{array}
\usepackage{makecell}
\usepackage{orcidlink}

\ifCLASSOPTIONcompsoc
  \usepackage[caption=false,font=normalsize,labelfont=sf,textfont=sf]{subfig}
\else
  \usepackage[caption=false,font=footnotesize]{subfig}
\fi

\usepackage{stfloats}
\usepackage{url}

\newcolumntype{D}{>{\centering\arraybackslash}c}

\begin{document}

\title{Multi-view 3D surface reconstruction from SAR images by inverse rendering}

\author{Emile~Barbier-{}-Renard\orcidlink{0000-0003-2967-6082},
        Florence~Tupin\orcidlink{0000-0002-3110-8183},~\IEEEmembership{Senior~Member~IEEE,}
        Nicolas~Trouvé
        and~Loïc~Denis\orcidlink{0000-0001-9216-8318},~\IEEEmembership{Senior~Member~IEEE}

\thanks{This work was supported in part by Agence de l'Innovation de Défense (AID) and in part by the Apprentissage statistique pour l’imagerie SAR multi-dimensionnelle (ASTRAL) Project under Grant ANR-21-ASTR-0011. \it{(Corresponding author: Emile Barbier-{}-Renard)}}
\thanks{The authors would like to thank ONERA and DGA for providing the EMPRISE\textsuperscript{\textregistered} simulator.}
\thanks{Emile Barbier-{}-Renard and Florence Tupin are with LTCI, Télécom Paris, Institut Polytechnique de Paris, 91120 Palaiseau, France (e-mail: emile.barbier.renard@telecom-paris.fr; florence.tupin@telecom-paris.fr).}
\thanks{Nicolas Trouvé is with the DEMR, ONERA, The French Aerospace Laboratory, 91761 Palaiseau, France (e-mail: nicolas.trouve@onera.fr).}
\thanks{Loïc Denis is with the Laboratoire Hubert Curien, UMR 5516, CNRS, Institut d'Optique Graduate School, Université de Lyon, UJM-Saint-Étienne, 42023 Saint-Etienne, France (e-mail: loic.denis@univ-st-etienne.fr).}
}


\maketitle

\begin{abstract}
3D reconstruction of a scene from Synthetic Aperture Radar (SAR) images mainly relies on interferometric measurements, which involve strict constraints on the acquisition process. These last years, progress in deep learning has significantly advanced 3D reconstruction from multiple views in optical imaging, mainly through reconstruction-by-synthesis approaches pioneered by Neural Radiance Fields. In this paper, we propose a new inverse rendering method for 3D reconstruction from unconstrained SAR images, drawing inspiration from optical approaches. First, we introduce a new simplified differentiable SAR rendering model, able to synthesize images from a digital elevation model and a radar backscattering coefficients map. Then, we introduce a coarse-to-fine strategy to train a Multi-Layer Perceptron (MLP) to fit the height and appearance of a given radar scene from a few SAR views. Finally, we demonstrate the surface reconstruction capabilities of our method on synthetic SAR images produced by ONERA's physically-based EMPRISE\textsuperscript{\textregistered} simulator. Our method showcases the potential of exploiting geometric disparities in SAR images and paves the way for multi-sensor data fusion.
\end{abstract}


\IEEEpeerreviewmaketitle

\section{Introduction}
\IEEEPARstart{S}{ynthetic} Aperture Radar (SAR) is an active remote sensing technique based on the measurement of echoes of a radar pulse backscattered by the scene. The content of SAR images is directly related to the geometry of the observed terrain, yet 3D reconstruction from a set of SAR images remains a complex task. The main methods for multi-image reconstruction in SAR (interferometry \cite{bamler_interferometry_1998}, tomography \cite{fornaro_tomography_2014}) rely on the phase shift between complex amplitudes collected at different antenna positions. It requires using very similar views so that phase changes can be solely attributed to differences in the path length due to geometry and not to changes of the complex backscattering coefficients. Other approaches are incoherent; they leverage only the intensity of the echoes by extracting and matching sparse spatial features such as bright lines (radargrammetry) \cite{palm_urban_radargrammetry_2012} or use a single view with a strong scene homogeneity assumption such as radarclinometry \cite{paquerault_radarclinometry_1996}. More recently, deep-learning-based single-image reconstruction for urban scenes has been demonstrated \cite{recla_sarsingle_2022}: a network is trained to infer the height map, projected in SAR geometry, from a single view. The network uses geometrical clues (in particular, shadows) which restricts its applicability to urban areas. Geometric disparities between SAR images are thus a promising, yet underused aspect for surface reconstruction.

In optical imaging, recent progress in neural networks has paved the way for a new generation of 3D reconstruction algorithms from multiple views. These approaches, which rely on the inversion of a physically-based rendering model and spearheaded by Neural Radiance Fields (NeRF) \cite{mildenhall_nerf_2020}, allow for the exploitation of unconstrained and sparse data to recover 3D information of a scene. The main idea is to fit a neural representation of the geometry and radiance of a scene by minimizing the dissimilarity between observed images and synthesized images generated according to the current 3D reconstruction. By using a differentiable rendering model, the gradient of this dissimilarity with respect to the network parameters can be computed and the parameters optimized. The rendering model is derived from ray casting volumic rendering, and the reconstruction models the geometry as a density field. The quality and efficiency of NeRF reconstructions rely significantly on a good positional encoding of 3D coordinates which many follow-up papers greatly improved such as \textit{Instant-NGP} \cite{mueller_instant_2022}.

While the initial works focused mainly on improving the fidelity of the novel views, many follow-up papers soon specialized the approach for different tasks such as implicit surface reconstruction by learning a signed distance field \cite{yariv_volume_2021,wang_neus_2021}. In the field of remote sensing, applications on passive sensors require the knowledge of the sun position to correctly predict the shadows as well as learning a different reflectivity for each image to explain the drastically different appearances from one view to another depending on the weather or the time of day \cite{mari_satnerf_2022,mari_eonerf_2023_CVPR}.

In the last year, contributions in active imaging have been submitted for various modalities. Regarding synthetic aperture acquisitions, the few SAS \cite{reed_volsas_2023} and SAR \cite{lei_SARNERF_2024, ehret_radarfields_2024} NeRF publications rely on the same equation of volume rendering popularized by the original NeRF article. Each ray is sampled regularly, once for each range cell, and backscattered intensities are accumulated across the rays. With SAR-NeRF \cite{lei_SARNERF_2024}, Lei et al. focused more on applying novel view synthesis of objects for data augmentation than on achieving the 3D reconstruction of a scene, and suggested learning for each voxel of a 3D matrix a density and a scattering intensity.

Meanwhile in \textit{Radar Fields} \cite{ehret_radarfields_2024} by Ehret et al., the geometry of the scene is modelled by a discrete elevation model which is turned into a density field using the transform introduced by VolSDF \cite{yariv_volume_2021} to use the framework of volume rendering. This method reports good surface reconstructions but has only been applied yet on synthetic data generated with its own rendering model. The use of a volume rendering approach is computationally costly since it requires sampling the whole volume.

In this paper, we propose a novel deep learning-based method for 3D reconstruction from unconstrained SAR images, drawing inspiration from these optical reconstruction by image synthesis approaches. Our method introduces a simplified differentiable rendering model that generates SAR images from a DSM and a radar backscattering coefficients map. The proposed model departs from the typical volume rendering used in most NeRF derived approaches by introducing a new rasterization of scatterers fully leveraging the DSM representation of the scene. Adaptations of volume rendering for active imaging must sample the volume both in angle and in range which is quite inefficient when considering surfaces, and insufficient sampling density leads to noisy renders. In opposition, our new model leverages the heightmap representation by selecting scatterers directly and exclusively on the surface and rasterizing piecewise-linear approximation of the terrain. We propose a coarse-to-fine strategy which reduces even more the sample count in the earlier phases of training by benefitting from the noise-free images our rendering model can synthesize from only a few samples. We demonstrate the surface reconstruction capabilities of our method by training a MLP to predict the terrain from only a few SAR images on synthetic SAR image datasets produced by ONERA’s physically based EMPRISE\textsuperscript{\textregistered} \cite{emprise_site,trouve_emprise_2024} simulator.

\section{SAR differentiable rendering model}

\subsection{SAR image formation model}

\begin{figure}
     \centering
    \subfloat[]{%
         \centering
         \includegraphics[width=0.48\linewidth]{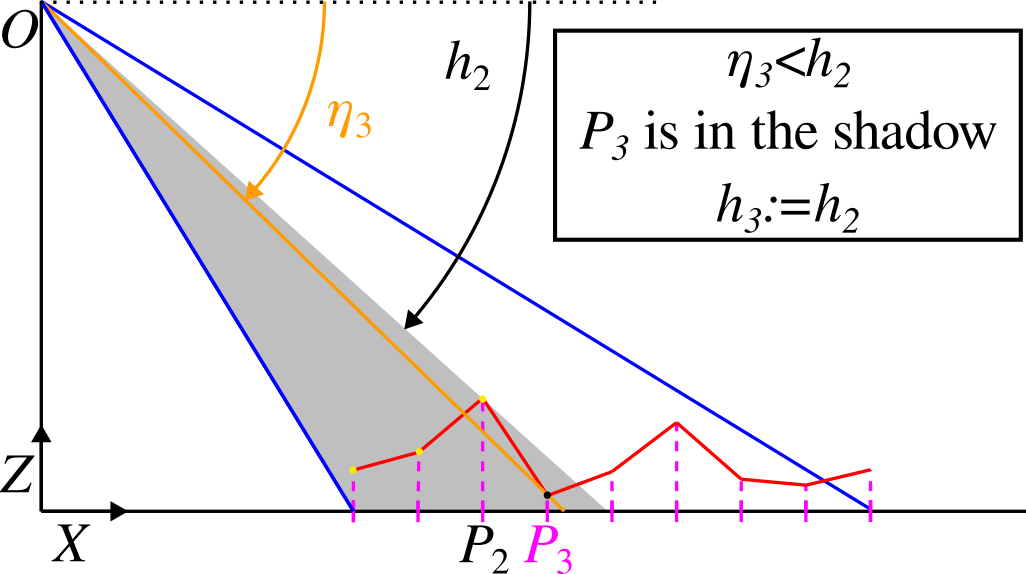}}
     \hfill
    \subfloat[]{%
         \centering
         \includegraphics[width=0.48\linewidth]{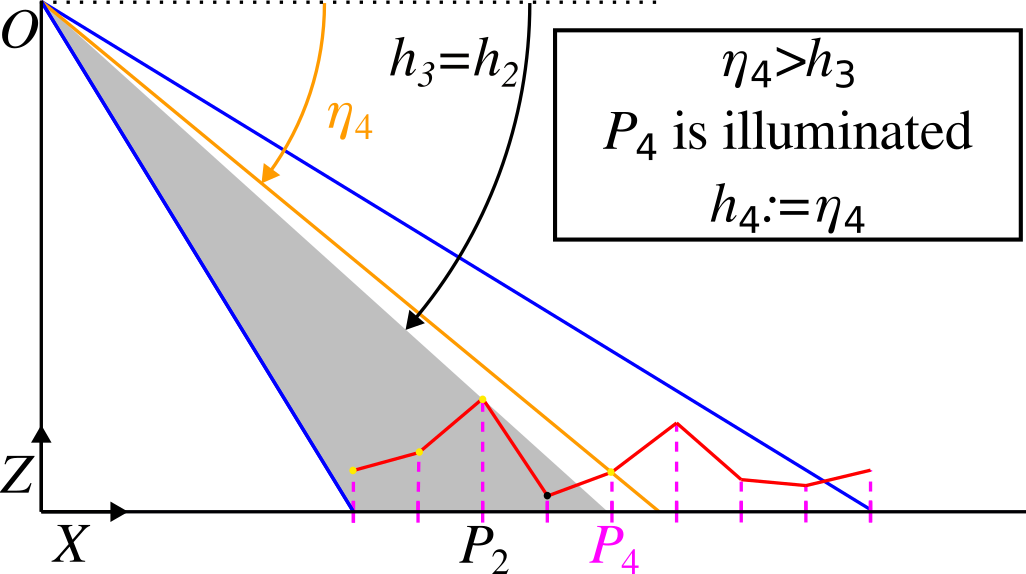}}
    \caption{Iterative determination of shadow areas:\\
            (a) Estimation of $v_3$, indicatrix of the shadow of sample $P_3$ knowing $h_2=\eta_2$, the negative slope between the antenna and $P_2$, the latest illuminated sample.\\
            (b) Estimation of $v_4$, indicatrix of the shadow of sample $P_4$ knowing  $h_3=\eta_2$ from the previous iteration.}
    \label{fig:shadows}
\end{figure}

Consider a stripmap acquisition in a radar band with negligible vegetation and ground penetration. The observed scene can be described by an elevation map $\mathcal Z(x,y)$, and its backscattering coefficients map $\mathcal B(x,y)$. When considering direct backscattering, the expected intensity in the $m$th range cell at a given azimuth in the processed image corresponds to the sum of the echoes backscattered by all the illuminated scatterers at the correct range, such that:
\begin{equation}
\label{eq_integral}
\bar I_m = \iint \sigma^0(x,y,\mathcal{Z}(x,y)) \cdot \mathcal{C}_m(x,y,\mathcal{Z}(x,y))\cdot dxdy
\end{equation}
Where $\sigma^0$ is the normalized radar cross section and $\mathcal{C}$ the indicatrix of contribution, equal to $1$ when a sample contributes to the considered cell. It is modelled by the product of:
\begin{enumerate}
    \item $\mathcal{U}(x,y,z)$, indicatrix of contribution to the azimuth row;
    \item $\mathcal{W}_m(x,y,z)$, indicatrix of contribution to the $m$th range cell;
    \item $\mathcal{V}(x,y,z)$, indicatrix of shadow reporting if the surface at $(x,y,z)$ is illuminated by the radar or in a shadow.
\end{enumerate}

\subsection{Discretization}

In order to efficiently evaluate the expected intensity, we propose a discretization of equation (\ref{eq_integral}) by decomposing the surface into small quadrilateral patches. Consider $K+1$ points $P_k$ sampled on the surface in the zero-doppler plane, uniformly in ground range. By joining them we form $K$ contiguous segments $[P_k,P_{k+1}]$ of respective lengths $l_k$ (red piecewise-linear curve in figure \ref{fig:shadows}). Stretching these segments along the azimuth axis by $\delta a$, the azimuth cell resolution, we construct $K$ rectangular surface patches. By definition, these patches contribute only to the azimuth row such that $\mathcal{U}(x,y)=1$ at each of their point. Writing $\hat v_k$ and $\hat w_{m,k}$ the shadow and range cell contribution terms associated to the $k$th patch, the estimated intensity corresponds to the summation of the contributions of the rectangular patches:
\begin{equation}
\label{eq_diff}
\hat I_m = \delta a\sum_{k=0}^{K-1}\sigma^0_k\cdot\hat v_k\cdot\hat w_{m,k}\cdot l_k
\end{equation}

Summing the contributions of rectangular patches provides an accurate estimation of the contributing area. When a coarse scattering rate is used, i.e. larger patches are considered, each patch contributes to several resolution cells which prevents creating gaps in the synthesized images. This last property is leveraged by the coarse-to-fine training strategy presented in \ref{ss:coarse_to_fine}.

\subsection{Shadows}

We consider the positions $P_k$ ordered by increasing ground range values, as illustrated figure \ref{fig:shadows}. The slope of the line $(O, P_k)$ between the antenna and the point is written $\eta_k$. The presence of shadow $v_k$ at location $P_k$ depends only on $P_j, j<k$, the location of the point illuminated last. $P_k$ is illuminated only if it is located above the shadow demarcation line $(O,P_j)$ of slope $h_k=\eta_j$. We introduce an iterative algorithm illustrated in figure \ref{fig:shadows} to update the presence of shadow $v_{k+1}$ and the slope if the shadow demarcation $h_{k+1}$ from $v_k$ et $h_k$. Setting $v_0 = 1$ et $h_0 = 0$:

\begin{equation}
\label{eq_vis}
\begin{cases}
    v_{k+1} = \mathcal{S}_\xi(\eta_k-h_k)\\
    h_{k+1} = \eta_k\cdot v_{k+1}+h_k\cdot (1-v_{k+1})
\end{cases}
\end{equation}

$\mathcal{S}_\xi$ is a logistic sigmoid of steepness $\xi$ approximating the unit step function and introducing differentiability in the calculation of $\hat v_k$, the shadow indicatrix associated to the $k$th rectangular patch. To simplify computations, we consider the scatterer $k$ illuminated if and only if its far extremity $P_{k+1}$ is illuminated, leading to $\hat v_k = v_{k+1}$.

\subsection{Contribution to radar cells}
The contribution of a patch to a radar cell $\hat w_{m,k}$ corresponds to the proportion of the segment $[P_k,P_{k+1}]$ belonging to the $m$th range cell. Thus we ensure the sum of the contributions of a scatterer across the cells does not exceed $1$.

We call $r_m^-$ and $r_m^+$ the lower and upper radius of the $m$th band and $d_k^-$ and $d_k^+$ the slant ranges of the extremities $P_k$ and $P_{k+1}$ of the segment. Supposing that the isochron lines are locally straight lines, the fraction of $[P_k,P_{k+1}]$ belonging to the $m$th range cell can be simply formulated with a projection on the line of sight as an operation on the intervals, using $\mathcal D([AB])$ the length of interval $[AB]$:
\begin{equation}
    w_{m,k}=\frac{\mathcal{D}([d_k^-d_k^+]\cap[r_m^-r_m^+])}{|d_k^+-d_k^-|}
\end{equation}

The contribution term is finally deduced using $M_\mu$, a smooth maximum function with an accuracy parameter $\mu$:
\begin{equation}
\label{eq_contrib}
\begin{split}
    \hat w_{m,k}=\frac{\mathcal{M}_\mu(d_k^-,r_m^+)+\mathcal{M}_\mu(d_k^+,r_m^-)}{d_k^+-d_k^-}\\-\frac{\mathcal{M}_\mu(d_k^+,r_m^+)+\mathcal{M}_\mu(d_k^-,r_m^-)}{d_k^+-d_k^-}
\end{split}
\end{equation}

Wher $\mathcal{M}_\mu$ is the a smooth maximum unit function \cite{biswas_smu_2022}:
\begin{equation}
\label{eq_smu}
    \mathcal{M}_\mu(a,b) = \frac{1}{2}\left(a + b + \frac{(a-b)^2}{\sqrt{(a-b)^2+\mu^2}}\right)
\end{equation}

\subsection{Normalized radar cross-section}

The normalized cross-section $\sigma^0_k$ of the patch associated to $[P_k,P_{k+1}]$ is estimated at its middle, at the point ${P_{k+\frac{1}{2}}=\frac{P_k+P_{k+1}}{2}}$ and considers a Lambertian surface defined by a backscattering coefficient and a cosine attenuation \cite{gelautz_alpine_1998} fitting our targeted incidence angle. Given the normal to the surface $\vec n_k$ deduced from the quadrilateral patch and the local incidence unit vector $\vec u_k$ pointing from the sensor in the direction of $P_{k+\frac{1}{2}}$:
\begin{equation}
\label{eq_sigma0}
\sigma^0_k(\vec u_k)=\mathcal{B}(P_{k+\frac{1}{2}})\cdot |\vec u_k \cdot \vec n_k|
\end{equation}

\section{Surface reconstruction strategy}

\subsection{Training protocol}\label{ss:neural_representation}

\begin{figure}[!t]
\centering
\includegraphics[width=2.5in]{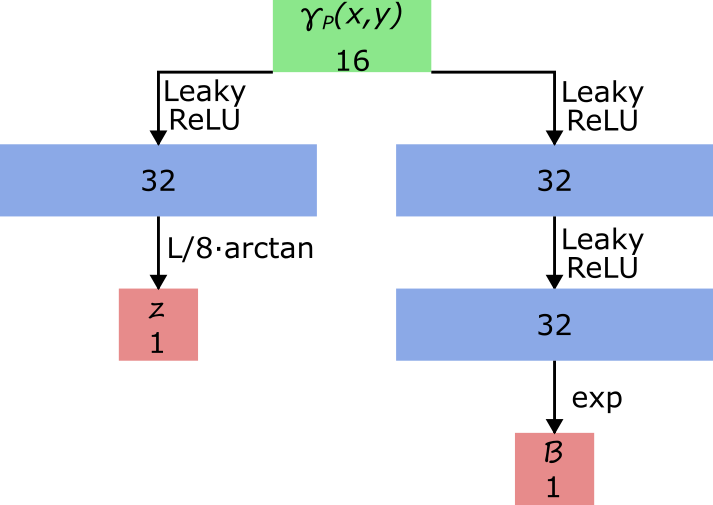}
\caption{Architecture of the fully-connected network. Input: $(x,y)$, a position on the ground, encoded with  is Instant-NGP's positional encoding $\gamma_P$  \cite{mueller_instant_2022}. Outputs: the elevation $\mathcal{Z}(x,y)$ and backscattering coefficient $\mathcal{B}(x,y)$.}
\label{fig_network}
\end{figure}

We model the elevation $\mathcal{Z}$ and backscattering coefficients map $\mathcal{B}$ within a bounding box of side $L$ using the MLP described fig \ref{fig_network}. It relies on Instant NGP's positional encoding $\gamma_P$ for the 2D position of the sample. The exponential activation applied to $\mathcal{B}$ ensures the prediction of positive values and the activation $\frac{L}{8}\cdot\arctan$ of the elevation model bounds its values within reasonable limits.

Because every range cell in each row rely on the same samples (only the contribution term of each sample $w_{m,k}$ is specific to a cell), it is far more efficient to render a row all at once rather than individual pixels. Therefore, given a collection of $N$ views of azimuth and range dimensions $N_A \times N_R$ and the associated acquisition parameters, we select and render for each minibatch a random subset of $M$ lines across all the images. In all our experiments, $M$ is greater than $80$ to ensure an adequate distribution of the samples on the surface at each minibatch while still meeting memory constraints. The ground ranges sampled are slightly shifted by a random value along their line of sight to avoid over-fitting at specific coordinates.

We fit the MLP by minimizing the loss $\mathcal{L}= \mathcal{L}_D+\lambda\mathcal{L}_{TV}$ using Adam optimizer.

$\mathcal{L}_{D}$ corresponds to the negative log-likelihood of the log-intensities, which takes into account the distribution of the speckle and allows for training on noisy images \cite{dalsasso_sar2sar_2021}. Writing $I$ the observed intensities and $\hat I$ the synthesized speckle-free intensities, it is defined as:
\begin{equation}
\label{eq_loss}
\mathcal{L}_D(\hat I) = \frac{1}{M\cdot N_R}\sum_{m=1}^{M}\sum_{n=1}^{N_R} \log\left(\frac{\hat I_{m,n}}{I_{m,n}}\right) + \frac{I_{m,n}}{\hat I_{m,n}}
\end{equation}

$\mathcal{L}_{TV}(\mathcal{B})$ is a total variation regularization over the predicted values of $\mathcal{B}$, used to penalize over-fitting on $\mathcal{B}$ and thus favour fitting the geometry. The weight $\lambda=0.1$ remains constant.

\subsection{Coarse to fine approach}\label{ss:coarse_to_fine}

For an efficient reconstruction of the multiple scales present in the geometry and to speed up the initial phases of the training process, we gradually increase the number of samples $K$ and thus the ability to represent high resolution features.

Given a target number of samples per line $K_f$ and a target accuracy $\mu_f$, we also define the initial subsampling factor $\beta_0$. Annealing $\beta$ from $\beta_0$ to $1$ gives the parameters $K=K_f/\beta$ and $\mu=\mu_f\cdot\beta$ for the current minibatch.

Furthermore, in order to regulate the frequency bias during earlier phases of training we apply Nerfies' windowing on frequency bands \cite{park_nerfies_2021, lin_barf_2021} on the output of the positional encoding $\gamma_P$. We dynamically determine for each sample $k$ of ground range $X_k$ the targetted maximum frequency level $l\in[0,L]$ such that it corresponds to $\frac{|X_{k+1}-X_{k-1}|}{2}$.

\section{Experiments}

In order to demonstrate the ability of our method to retrieve DSMs from only a few SAR images, we apply the described surface reconstruction method to synthetic SAR data generated by the physically-based EMPRISE\textsuperscript{\textregistered} simulation software \cite{emprise_site} developed by ONERA under DGA's funding. The reconstructions are evaluated quantitatively by computing the RMSE on the recovered elevations (restricted to emerged lands and to the areas of overlap of at least 2 images), the error values are shown in table \ref{table_results}.

\begin{figure*}[tbh]
     \centering
     \begin{tabular}{@{}
      c @{\hspace{12pt}}
      c @{\hspace{24pt}}
      c @{\hspace{24pt}}
      c @{\hspace{12pt}}
      c
    }
    & \textbf{Reference}
    & \makecell{\textbf{Reconstruction}\\\textbf{from 5 images}}
    & \makecell{\textbf{Reconstruction}\\\textbf{from 2 images}} & \\

    \rotatebox[origin=c]{90}{\textbf{DSM}}
    & \raisebox{-0.5\height}{\includegraphics[height=4cm, trim={1.9cm 0.8cm 2.3cm 0},clip]{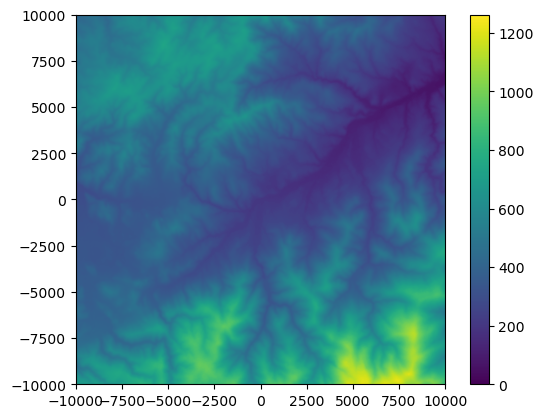}}
    & \raisebox{-0.5\height}{\includegraphics[height=4cm, trim={1.9cm 0.8cm 2.3cm 0},clip]{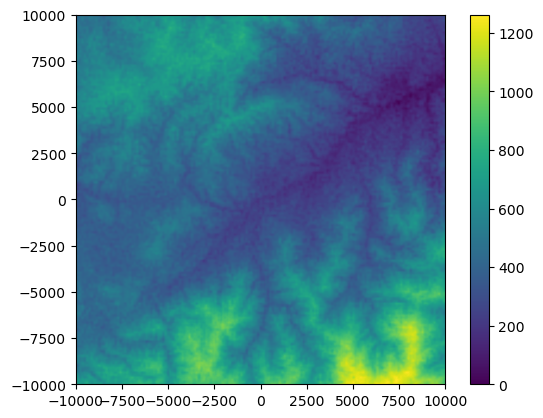}}
    & \raisebox{-0.5\height}{\includegraphics[height=4cm, trim={1.9cm 0.8cm 2.3cm 0},clip]{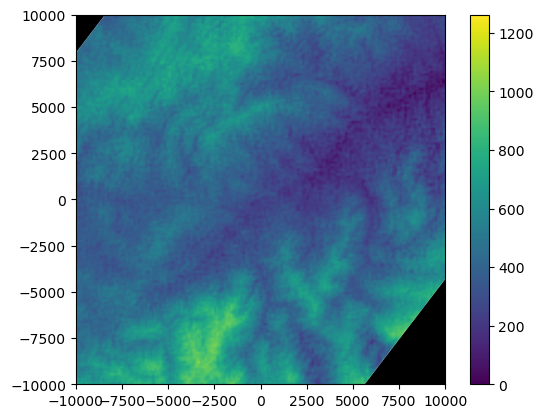}}
    & \raisebox{-0.5\height}{\includegraphics[height=4cm, trim={11.9cm 0.7cm 0 0},clip]{fig/results/Forez_2/pred_dem_masked.png}}\\

    \rotatebox[origin=c]{90}{\textbf{Log-intensity SAR image}}
    & \raisebox{-0.5\height}{\includegraphics[height=4.6cm, trim={1.15cm 0.8cm 1.9cm 0},clip]{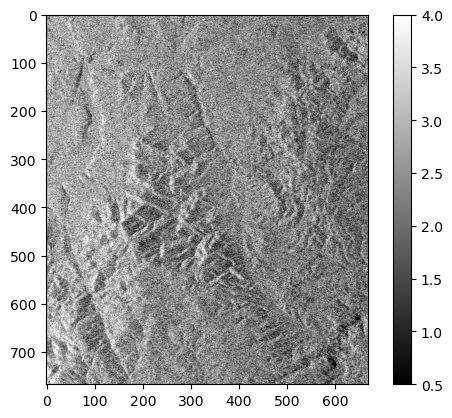}}
    & \raisebox{-0.5\height}{\includegraphics[height=4.6cm, trim={1.15cm 0.8cm 1.9cm 0},clip]{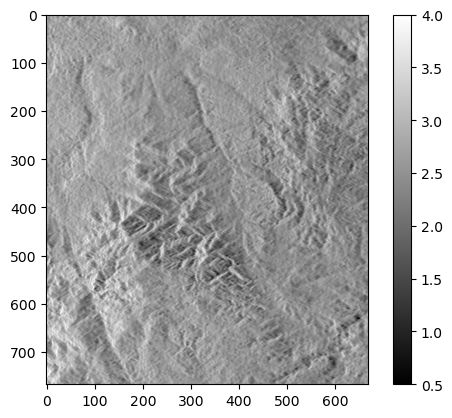}}
    & \raisebox{-0.5\height}{\includegraphics[height=4.6cm, trim={1.15cm 0.8cm 1.9cm 0},clip]{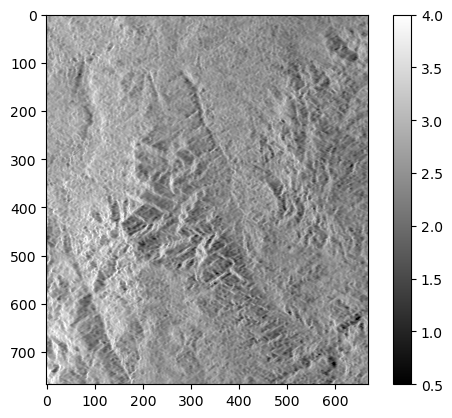}}
    & \raisebox{-0.5\height}{\includegraphics[height=4.6cm, trim={9.97cm 0.65cm 0 0},clip]{fig/results/Forez_2/pred_sar.png}}\\

    & EMPRISE\textsuperscript{\textregistered}
    & Our model
    & Our model
    & 
    
    \end{tabular}
    \caption{Results of the DSM reconstruction on the low resolution Forez scene. Areas not visible in both images are masked out on the DSM reconstructed from ascending/descending passes (third column).}
    \label{fig:results forez}
\end{figure*}

The first reconstruction is carried out on a set of 5 low-resolution views of the Forez, France, using NASA's SRTM mission digital elevation model \cite{SRTMGL1} and with uniform material properties over the whole area. The second reconstruction is performed on the same scene using only ascending and descending passes. A sample of the EMPRISE\textsuperscript{\textregistered} simulation as well as the results of the reconstruction are shown in figure \ref{fig:results forez}. These results illustrate the ability to reconstruct a digital surface model using our inverse rendering approach, even for a limited number of views. Reconstruction only from ascending/descending passes shows a degradation of the RMSE, but still provides a fidelity finer than the source SAR images' pixel resolution. This higher RMSE comes from a grainier reconstructed heightmap due to a slight over-fitting to the speckle: fewer views also translate to a reduction in available speckle realisations at each spatial position.

\begin{figure}[tbh]
     \centering
     \begin{tabular}{@{}
      c @{\hspace{3pt}}
      c @{\hspace{3pt}}
      c @{\hspace{0pt}}
      c
    }
    & \textbf{Reference}
    & \makecell{\textbf{Reconstruction}\\\textbf{from 5 images}} & \\

    \rotatebox[origin=c]{90}{\textbf{DSM masked for water}}
    & \raisebox{-0.5\height}{\includegraphics[height=4cm, trim={1.9cm 0.8cm 2.3cm 0},clip]{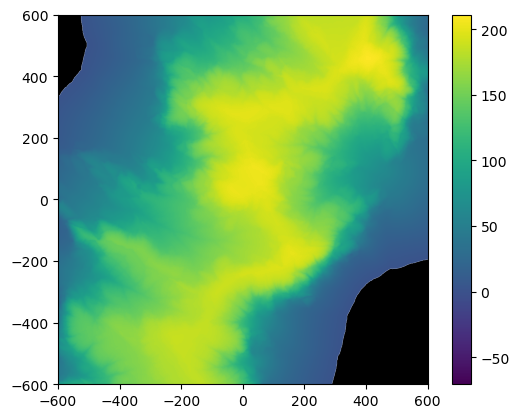}}
    & \raisebox{-0.5\height}{\includegraphics[height=4cm, trim={1.9cm 0.8cm 2.3cm 0},clip]{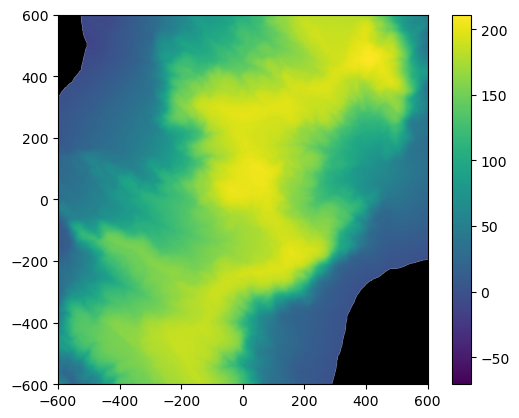}}
    & \raisebox{-0.5\height}{\includegraphics[height=4cm, trim={11.3cm 0.7cm 0 0},clip]{fig/results/Island_5/ref_dem.png}}\\

    \rotatebox[origin=c]{90}{\textbf{Scattering coefficients}}
    & \raisebox{-0.5\height}{\includegraphics[height=4cm, trim={1.9cm 0.8cm 2.3cm 0},clip]{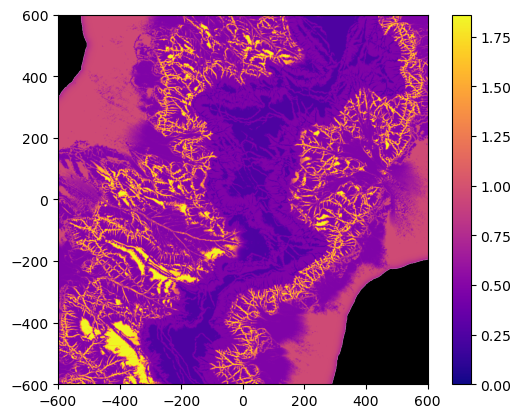}}
    & \raisebox{-0.5\height}{\includegraphics[height=4cm, trim={1.9cm 0.8cm 2.3cm 0},clip]{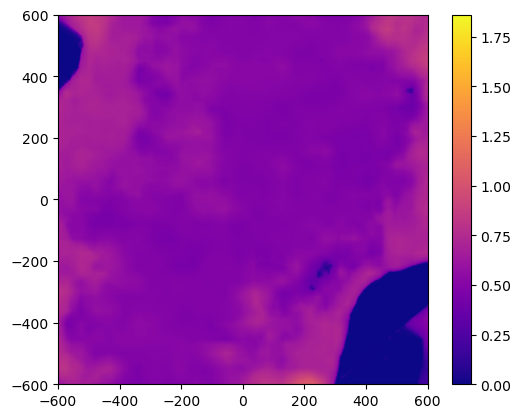}}
    & \raisebox{-0.5\height}{\includegraphics[height=4cm, trim={11.3cm 0.7cm 0 0},clip]{fig/results/Island_5/ref_appearance.png}}\\
    
    \rotatebox[origin=c]{90}{\textbf{Log-intensity SAR image}}
    & \raisebox{-0.5\height}{\includegraphics[height=2.6cm, trim={1.15cm 1.6cm 1.9cm 1.5cm},clip]{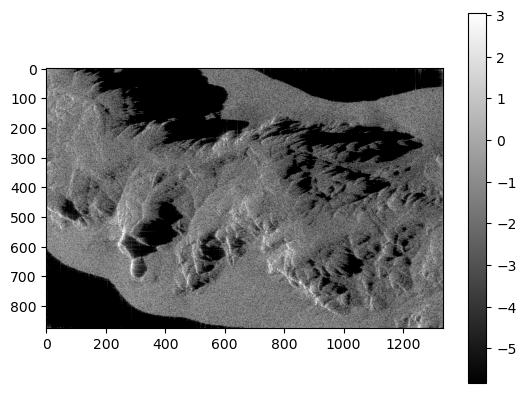}}
    & \raisebox{-0.5\height}{\includegraphics[height=2.6cm, trim={1.15cm 1.6cm 1.9cm 1.5cm},clip]{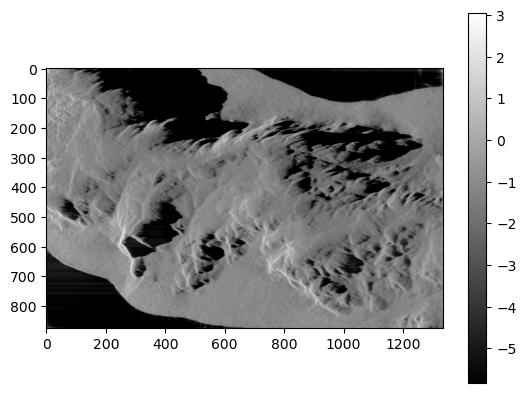}}
    & \raisebox{-0.5\height}{\includegraphics[height=3.9cm, trim={11.85cm 0.7cm 0 0},clip]{fig/results/Island_5/ref_sar.png}}\\

    & EMPRISE\textsuperscript{\textregistered}
    & Our model
    & 
    
    \end{tabular}
    \caption{Results of the surface learning on the high resolution island scene with multiple materials. Reference scattering coefficients are estimated from the mangia model by an evaluation with 45° incidence.}
    \label{fig:results islands}
\end{figure}

The third experiment is conducted on a set of 5 high-resolution views of an island surrounded by water, with multiple materials present in the scene. The EMPRISE\textsuperscript{\textregistered} engine describes the $\sigma^0$ of different materials using the mangia model \cite{houret_emprisecalibration_2023}, which is more complex than our cosine hypothesis. Despite facing multiple materials, the results presented in figure \ref{fig:results islands} demonstrate the ability of our approach to separate geometry and appearance: the computed RMSE on the elevation confirms the visual quality of the terrain reconstruction, and our simple backscattering representation proves sufficient to operate well even with a limited number of views. Furthermore, the presence of cliffs in the scene leads to the formation of large shadow areas with an aspect similar to the submerged areas, yet the predicted backscattering coefficients show no confusion between the submerged areas (reconstructed with a low value as expected) and the areas near the cliffs that appear in the shadow in some of the views. Although our reconstruction does not retrieve a precise map of the materials, their demarcation can be observed on flatter terrain, and lost texture details are not misinterpreted as geometric features.

\begin{table}[!t]
\renewcommand{\arraystretch}{1.3}
\caption{Quantitative evaluation of the reconstructed DSMs}
\label{table_results}
\centering
\begin{tabular}{|c|c|c|c|}
\hline
Scene & \multicolumn{2}{c|}{Forez} & Island\\
\hline
\hline
Cell resolution & \multicolumn{2}{c|}{75m} & 0.75m \\
\hline
Material properties & \multicolumn{2}{c|}{Uniform} & Multiple \\
\hline
Number of views & 5 & 2 & 5 \\
\hline
RSME $\downarrow$ & 36.7m & 52.9m & 4.72m\\
\hline
\end{tabular}
\end{table}

\section{Conclusion}

In this paper, we introduced a simplified differentiable SAR intensity rendering model, adapted to the surface reconstruction by synthesis approach. This new physically-based model considers the properties of the heightmap representation of a scene to render intensity images from fewer samples than competing approaches relying on volume rendering by reducing the sampling degree of freedom from $3$ (through the whole volume) to $2$ (only on the surface). Compared to a density field, this representation is also more constrained, leading to faster and easier reconstruction of surfaces. Paired with a custom coarse-to-fine strategy, we demonstrated its ability to successfully retrieve a neural elevation model from only a few images and correctly separate geometry from texture. To go further than \textit{Radar Fields} \cite{ehret_radarfields_2024} we also establish our method's results beyond self-inversion through experiments on EMPRISE\textsuperscript{\textregistered} simulations, a step towards reconstructions from real SAR data.

\ifCLASSOPTIONcaptionsoff
  \newpage
\fi

\nocite{*} 

\bibliographystyle{IEEEtran}
\bibliography{bibtex/bib/IEEEabrv,bibtex/bib/activenerf,bibtex/bib/nerf,bibtex/bib/sar,bibtex/bib/utils,bibtex/bib/emprise}

\end{document}